\documentclass[10pt,twocolumn,letterpaper]{article}

\usepackage{iccv}
\usepackage{times}
\usepackage{epsfig}
\usepackage{graphicx}
\usepackage{amsmath}
\usepackage{amssymb}
\usepackage{bm}
\usepackage{booktabs}
\usepackage{tabularx}
\usepackage{microtype}
\usepackage{tikz}
\usepackage{adjustbox}
\usepackage{csquotes}
\usepackage{siunitx}
\usepackage{authblk}
\usepackage[light]{FiraSans} 
\usepackage[T1]{fontenc}
\usetikzlibrary{positioning, arrows.meta, calc, fit, backgrounds, matrix, fadings}
\usepackage[round,numbers]{natbib}

\usepackage[inline]{enumitem}

\usepackage[breaklinks=true,bookmarks=false]{hyperref}

\iccvfinalcopy 

\makeatletter
\renewcommand\AB@affilsepx{, \protect\Affilfont}
\makeatother

\begin{document}

\title{Can we predict the Most Replayed data of video streaming platforms?}

\author[1]{Alessandro Duico}
\author[1,2]{Ombretta Strafforello}
\author[1]{Jan van Gemert}
\affil[1]{Delft University of Technology} 
\makeatletter
\renewcommand\AB@affilsepx{\\ \protect\Affilfont}
\makeatother

\affil[2]{TNO}

\affil[ ]{\textit{\{a.duico, o.strafforello, j.c.vangemert\}@tudelft.nl}}

\renewcommand\Authands{, }

\maketitle
\ificcvfinal\thispagestyle{empty}\fi

\begin{abstract}
Predicting which specific parts of a video users will 
replay is important for several applications, including targeted advertisement placement on video platforms and assisting video creators.
In this work, we explore whether it is possible to predict the \emph{Most Replayed} (MR) data from YouTube videos. 
To this end, we curate a large video benchmark, the \emph{YTMR500} dataset, which comprises 500 YouTube videos with MR data annotations. 
We evaluate Deep Learning (DL) models of varying complexity on our dataset and perform an extensive ablation study. In addition, we conduct a user study to estimate the human performance on MR data prediction.
Our results show that, although by a narrow margin, all the evaluated DL models outperform random predictions. 
Additionally, they exceed human-level accuracy. This suggests that predicting the MR data is a difficult task that can be enhanced through the assistance of DL. 
Finally, we believe that DL performance on MR data prediction can be further improved, for example, by using multi-modal learning. 
%
%
We encourage the research community to use our benchmark dataset to further investigate automatic MR data prediction.
\end{abstract}

\vspace{-0.5cm}

\section{Introduction}
Video streaming has emerged as a dominant mode of online communication, representing 73\% of all internet traffic in 2017 \cite{Cisco}, 
with YouTube leading the way as the most popular platform.
Video streaming platforms accumulate, in addition to video data, a substantial amount of metadata, pertaining to users' watching habits and interests.
Notably, in May 2022 YouTube released a new feature that shows a line chart of the most frequently replayed moments of each video, the \emph{Most Replayed} data.
In addition to aiding YouTube users during video playback, this data can serve various other potential applications, such as optimizing advertisement placement and giving feedback to content creators -- for instance, suggesting uninteresting scenes to remove.
In both cases, it is desirable to predict the Most Replayed data \emph{before} publishing a video. For advertisers, it enables placing advertisements optimally from the very first views, thereby maximizing profits. For content creators, it allows cutting the video appropriately before it reaches the audience, preventing the reputational damage caused by re-uploading a video after collecting the data.

In this work, we investigate whether it is possible to predict the Most Replayed data using Deep Learning (DL). 
To this end, we collect YTMR500, a dataset of 500 vlog and travel videos with their corresponding Most Replayed data and pre-extracted video features. For a comprehensive description of our dataset collection process, readers are referred to the Supplementary Material (Section \ref{section:supplementary_ytmr500}). 
We evaluate two DL models on YTMR500, consisting in a fully-connected network and an attention-based architecture, inspired by the PGL-SUM model that Apostolidis \etal~\cite{PglSum} proposed for video summarization.
%
We compare the results of DL against human performance, which we estimate through a crowdsourced user study.

We make the following contributions: (1) We introduce \textit{YTMR500}, a dataset of 500 videos and the corresponding Most Replayed data, that can foster research on most replayed data predictions in videos; (2) We design a variant of the PGL-SUM~\cite{PglSum} architecture, adapted to predict the Most Replayed data for unseen videos; (3) We perform a user study to evaluate human performance on MR data prediction. Our results show that predicting the Most Replayed data is challenging for human annotators and that our model surpasses human performance.
%
The YTMR500 dataset and our code are publicly available\footnote{\url{https://github.com/ombretta/most-replayed-data}}. 




\section{Predicting the Most Replayed data in a video}
\subsection{Problem statement}
Given a sequence of segments $\bm{V}$ in a video, we design a model to learn a function $F$ that maps $\bm{V}$ into the Most Replayed (MR) data $Y$.
Specifically, the input is a sequence of 1024-dimensional video features, $\bm{v_t}$, and the output is a sequence of scores $y \in [0,1]$. Formally,

\begin{align}
\bm{V} &=\{\bm{v_t}\}_0^T, \bm{v_t} \in \mathbb{R}^{N=1024} \\
Y      &= \{y_i\}_0^I, I=100, y_i \in [0,1] \\
F &: \bm{V} \to Y.
\end{align}

We want the model to determine the relative MR data of video segments when compared to one another, while we do not care about predicting the exact value of the Most Replayed data. Thus, instead of training the model using a Mean Squared Error loss, as in \cite{PglSum}, we opt for a ranking loss, namely PyTorch's \href{https://pytorch.org/docs/2.0/generated/torch.nn.MarginRankingLoss.html}{\texttt{MarginRankingLoss}}.
\begin{equation}
\label{equation:marginrankloss}
        \mathcal{L}( \hat{y}_i,\hat{y}_j,s) = \max(0, {-} s_{i,j}\cdot(\hat{y}_i {-} \hat{y}_j)+\text{margin})
\end{equation}

To match this setup, we construct a ranking of video segments based on the ground-truth (GT) $y_i$ and predicted $\hat{y}_i$ MR data scores. The \texttt{MarginRankingLoss} forces the model to predict MR scores that result in a ranking as close as possible to the GT. 
Concretely, the loss is applied to the video segments in a pairwise fashion. In Equation~\ref{equation:marginrankloss}, the targets $s_{i,j}$ must be one of $\{1, -1, 0\}$. Here, $1$ indicates $y_i > y_j$, $-1$ indicates $y_i < y_j$, and $0$ indicates $y_i = y_j$.
To obtain these targets $s_{i,j}$, during training, we generate a comparison matrix of size $T\times T$ given by $S_{i,j} = \text{sgn}(y_i - y_j)$.
Therefore, the ranking of each pair of predictions $\{\hat{y}_i, \hat{y}_j\}$ is encouraged to match the ranking of the GT $\{y_i, y_j\}$.

\paragraph{Interpolation}
While the Most Replayed data in the YTMR500 dataset have fixed length, the number of video features varies with the duration of the videos. To use a recurrent or attention-based model, the length of the inputs and that of the outputs must match. We overcome this issue in two different ways: 
\begin{enumerate}
    \item \label{case:interpolate} interpolating the ground truth to match the size of the frame features;
    \item \label{case:bin_avg} computing a binned average of the frame features, to match the size of the ground truth.
\end{enumerate}

In Case \ref{case:interpolate}, the GT $Y$ is interpolated from its fixed size $100$ into a variable size $T$. As a result, the interpolated GT $\widetilde{Y}$ has a different size for each of the videos, depending on the video duration.
Considering that the number of frame features $T$ is always greater than $100$, this type of interpolation allows us to supply a larger amount of input data to the model, as compared to case 2.
Since the number of comparisons in the matrix $S$ grows quadratically with the size of the input, during training we randomly sample a subset of 10,000 comparisons, for each iteration and each video. Furthermore, the random sampling ensures that each video has the same contribution to the loss, independently of $T$. 

In Case \ref{case:bin_avg}, we divide the frame features $\bm{V}$ in $100$ bins of uniform size and compute the average of the features within each bin. Therefore, the binned frame features $\bm{\widetilde{V}}$ have a constant cardinality of $100$, regardless of the video length.
Using this strategy, the training becomes computationally easier, as there are fewer elements to be ranked.

\subsection{DL models} \label{section:model}

\paragraph{Fully connected model} As a baseline, we use a fully-connected model with two linear layers. We include a Dropout layer to mitigate overfitting on the training set. The final layer uses a Sigmoid activation function to coalesce the outputs into the $[0,1]$ range, similarly to the GT. 
It is worth noting that this architecture does not capture any temporal relationships between different segments of the video.

\paragraph{Attention-based model} \label{para:model-attention}
We investigate whether considering the temporal relationships across segments improves the MR prediction accuracy beyond our baseline.
We use an attention-based model, following the architecture of PGL-SUM (Apostolidis \etal~\cite{PglSum}) as closely as possible. 
PGL-SUM is based on Multi-Head Attention (MAH) performed globally, \ie, for the whole duration of the video, and locally, on a number of time windows -- 4 in our setup -- obtained from a partitioning of the video.
In the global attention module, we use 8 parallel attention heads, similarly to \cite{PglSum} and to the original Transformer \cite{vaswani2017attention}. In the local attention modules we use 4 heads.
An illustration of the deployed models is provided in the Supplementary Material (Section \ref{section:supplementary_method}).

\subsection{User study}
Since there exists no prior work on predicting the most replayed data in a video, there are no indications on the feasibility and difficulty of the task. We perform a user study to gain insights on how difficult predicting the most replayed data is for humans.
Our user study contains 30 videos randomly selected from the test set.
For the MR data prediction, we use a different setup than for the DL model. We do this because the task performed by the DL model, \ie, constructing a ranking of 100 video segments, is too complex for human annotators. 
We simplify the task by subdividing each video in 10 shots rather than 100 
and averaging the underlying 100 GTs into 10 bins.

We do not ask users to manually fill in the Most Replayed data, to prevent the influence of biases, \eg, a bias towards continuity of the score across segments.
Instead, we show a series of side-by-side comparisons of two video segments, to guide the users towards building a ranking.
The user study is composed of 
an introduction about the purpose of the Most Replayed data; the full video, sped up to $30$s for convenience; 19 pairwise comparisons, with the addition of an attention check.
Each comparison presents participants with two video shots, sped up to $10$s, along with the instructions: 
``\textit{Guess which of the two video shots has greater `Most replayed' score.}''
followed by the mutually exclusive options: ``Left'', ``Right'', and ``CONTROL''.
As part of the attention check, in one extra comparison, we place a video with a text overlay asking to choose the ``CONTROL'' option. Participants who fail this simple check are rejected.

The indices for the binary comparison were derived from the execution of the MergeSort algorithm \cite{knuth1998sorting} on 10 elements. 
The number of comparisons, 19, corresponds to the number of operations that the MergeSort algorithm needs to construct a total ordering.
The indices of the segments involved in the comparisons are randomly permuted for each user, so that any imbalance with the MergeSort indices is not reflected in the outcome.
Once we have obtained the answers to the pairwise comparisons, we construct a graph of the ordering and perform depth-first graph traversal to reconstruct a unique ranking of the segments. 

The user study was crowdsourced to approximately 300 paid workers on Amazon MTurk \cite{mturk}. 
Our crowdworkers population corresponds to the average demographics on Amazon MTurk, with a uniform distribution across genders, mainly comprising residents from the US and India, born between 1990 and 2000, as reported by mturk-tracker \cite{difallah2018demographics}.
The number of users assigned to each video was 10 to 11, which is sufficient to obtain a statistically significant result, according to Carvalho \etal~\cite{Carvalho_Dimitrov_Larson_2016how_many_crowdsourced}.

\section{Results}
\paragraph{Model training}

We use 5-fold cross validation, with a 80/20 training/test ratio. Therefore, of the 500 videos, 400 are utilized for training, and 100 for testing.
To train our models we follow a similar procedure to Apostolidis \etal~\cite{PglSum} and Fajtl \etal~\cite{fajtl2019vasnet}.
We use the Adam optimizer \cite{kingma2017adam} with learning rate $lr =$ \SI{5e-5} and L2 regularization $\lambda =$ \SI{1e-5} ..
Each batch contains only one sample, which is an entire video. This explains the low learning rate. 
For the \texttt{MarginRankingLoss}, we always use a margin of $0.01$, except for when we are training on 10 video shots, in which case we use a margin of $0.05$.
We train for 300 epochs, because at that point the training set accuracy reaches a plateau. Even though other research commonly picks the best epoch with respect to the test set \cite {PglSum, fajtl2019vasnet}, we refrain from doing so not to artificially boost our results.
Therefore, when reporting the scores we average over the last 50 epochs (250 to 299) and all 5 splits.

\paragraph{Evaluation metrics}
To evaluate our model, a ranking correlation metric could be used, \eg, Kendall's $\tau$~\cite{kendall_1938}. However, this would penalize equally errors at the bottom and at the top of the ranking. 
Furthermore, we prioritize the global ordering rather than the exact position of each element. For instance, permutations of adjacent segments in the predicted ranking should not be heavily penalized.

Hence, we use precision@K, a metric inspired from information retrieval \cite{schutze2008introduction}. Precision@K measures how many of the top K results are true positives, divided by K. Given that we do not work with binary labels, we classify the top K video shots in the ground truth as positives and the rest as negatives. 
Using these labels, the metric corresponds to the proportion of the top K video shots of the predicted ranking that are among the top K video shots in the ground truth.
We report precision@K for K in $\{15,30,50\}$, after interpolating the total number of video shots to 100, when required.
The selection of $K=15$ is inspired by the evaluation practices for video summarization in the literature \cite{apostolidis2021survey} which typically adhere to a portion of 15\% of the total duration.  
The values of $K=30$ and $K=50$ were chosen to assess precision at varying ranking depths.
%
In the context of the user study, given that there are only 10 video shots, we use values of K in $\{1,3,5\}$. It is worth noting that precision@1 corresponds to top-1 accuracy, a metric commonly used in image classification \cite{NIPS2012_c399862d}.

\subsection{DL models}
\label{subsection:results-model}
All models are able to sufficiently fit the training set, obtaining precision@50 above $80\%$ on the test set. However, the performance at test time is only marginally better than random.
Results at test time are shown at the top of Table \ref{table:ablation}.
It is surprising that our fully-connected baseline exhibits a satisfactory performance on this task, and the gains of the more complex PGL-SUM architecture are minimal.

We perform an ablation study on the full model, to understand the contribution of each component. At the bottom of Table \ref{table:ablation} we report the scores for our model without local attention, without global attention and without the residual connection. We only display Case \ref{case:interpolate} of the interpolation, since the two cases are almost identical in performance.
We discover that the removal of global or local attention does not heavily impact performance, contrary to the removal of the residual connection (shown in Supplementary Material, Figure 2). Therefore, we deduce that the crucial part of the learning is occurring on the input frame features, within the fully-connected layers at the end of the pipeline.

\subsection{User study}
\label{subsection:results-user-study}
To measure inter-rater agreement, we compute the Krippendorff's $\alpha$ \cite{krippendorff2004content} among all the users, for each video. Since the average $\alpha$ is $-0.017 \pm 0.009$, we conclude that the users' answers are generally not coherent with one another.
In Table \ref{table:user_study} we compare the precision@K of the users' rankings and those generated by the DL model, in the simplified scenario with 10 video segments.
The top section of the table shows the results computed on the 30 videos included in the user study, while the bottom section shows the results on the complete test set, averaged on 5 splits.
Note that for the YTMR500 test set, the standard deviation is computed between the results of the 5 splits only, not across all the videos in the test set.

We report scores for our attention-based model in two scenarios: firstly, when it is trained on the ground truth interpolated to 10 data points (``trained on 10'' in Table \ref{table:user_study}), to closely match the task given to the users; secondly, when it is trained on the complete ground truth (``trained on 100'' in Table \ref{table:user_study}), and averaged into 10 bins afterward.
Naturally, training on more data, in the ``trained on 100'' case, yields better performance.
Users are not able to perform significantly better than random on this task. 
Note that users do not undergo the same training as our DL models, which means they must base their predictions on their prior knowledge.
From the results of our user study, we believe that predicting the Most Replayed data from video segments is a difficult task for humans.



\begin{table}[tb]
\begin{tabularx}{0.5\textwidth}{@{}Xrrr@{}} \toprule
    Model                     & \small{prec.@1 (\%)} & \small{prec.@3 (\%)} & \small{prec.@5 (\%)}\\ 

    \midrule
    Random                    &  10 &  30 &  50 \\  
    \midrule
    \multicolumn{4}{c}{\textbf{User study test set} (30 videos)} \\
    \midrule
    Users (avg.)              & 9.6 $\pm$ 25.9 & 31.8 $\pm$ 20.4 & 51.3 $\pm$ 15.3 \\
    PGL-SUM trained on 10     &  $\mathbf{18.5 \pm 15.0}$ & 37.0	 $\pm$  10.4 & 53.0	 $\pm$  6.9 \\

    PGL-SUM trained on 100 & 17.5 $\pm$ 12.4 & $\mathbf{39.2 \pm 12.6}$ & $\mathbf{58.5  \pm 	9.5}$ \\
    \midrule
    \multicolumn{4}{c}{\textbf{YTMR500 test set} ($5$ splits $\times100$ videos)} \\
    \midrule
    PGL-SUM trained on 10      & 14.3 $\pm$ 2.9 & 35.0 $\pm$ 1.5 & 54.7 $\pm$ 1.5 \\
    PGL-SUM trained on 100   & $\mathbf{17.5 \pm 2.7}$ & $\mathbf{38.1 \pm 2.2}$ & $\mathbf{56.5 \pm 1.8}$ \\
    \bottomrule
    
\end{tabularx}
\caption{Results of our user study compared to our best DL model.
Users are not significantly better than random.
The DL models are always superior, and perform better when trained on more data, with one exception.
N.b. precision@$\{1,3,5\}$ are computed on rankings of 10 segments. The precision of the DL model is averaged over the last 50 epochs. Standard deviation is computed among the 30 videos for the user study test set and among the splits of a 5-fold cross validation for the YTMR500 test set.}
\label{table:user_study}
\end{table}



\begin{table}[tb]
\begin{tabularx}{0.5\textwidth}{@{}Xrrr@{}} \toprule
    Model           & prec.@15 & prec.@30 & prec.@50 \\ 
                    & (\%) & (\%) & (\%) \\ 
    \midrule
    Random          & 15 & 30 & 50 \\
    Fully-connected~\ref{case:interpolate}        & 21.5 $\pm$ 1.8 & 37.6 $\pm$ 0.9 & 56.3 $\pm$ 0.4 \\
    Fully-connected~\ref{case:bin_avg}    & 21.3 $\pm$ 1.6 & 37.4 $\pm$ 1.1 & 56.2 $\pm$ 0.7 \\
    PGL-SUM~\ref{case:interpolate}           &  $\mathbf{22.0 \pm 1.6}$  & $\mathbf{37.9 \pm 1.0}$ & $\mathbf{56.8 \pm 0.6}$ \\
    PGL-SUM~\ref{case:bin_avg}               & $22.0 \pm 1.9$ & $37.7 \pm 0.8$  & $56.4 \pm 0.6$ \\
    PGL-SUM~\ref{case:interpolate} w/o local attention & $20.6 \pm 2.0$ & $36.4 \pm 1.0$ & $55.6 \pm 1.0$ \\
    PGL-SUM~\ref{case:interpolate} w/o global attention & $21.5 \pm 1.7$ & $37.5 \pm 1.1$ & $56.6 \pm 0.7$ \\ 
    PGL-SUM~\ref{case:interpolate} w/o residual & $19.4 \pm 1.2$ & $35.6 \pm 1.5$ & $55.1 \pm 0.9$ \\ 
    \bottomrule
\end{tabularx}
\caption{Results on the test set for our models, followed by some ablations. 
All DL models perform better than random.
Surprisingly, the gains of the more complex PGL-SUM architecture are minimal, suggesting that attention between the segments is not fundamental.\\
N.b. precision@$\{15,30,50\}$ are computed on rankings of 100 segments. The numbers 1 and 2 refer to the type of interpolation: Case \ref{case:interpolate} is when the ground truth is interpolated and Case \ref{case:bin_avg} is when the frame features are averaged.}
\label{table:ablation}
\end{table}

\subsection{Discussion}
\label{subsection:discussion}

Based on our experiments in Section \ref{subsection:results-model}, we observe that using a more complex architecture does not induce a significant performance gain over the fully-connected baseline. Contrarily to our speculations, it seems that providing each segment with context about the full video does not improve the accuracy.
We hypothesize that the pre-extracted video features alone provide sufficient abstraction to the fully-connected network to generate a satisfactory output from individual segments. As demonstrated by the ablation without a residual connection, hiding the input features from the fully-connected layers hinders performance. 
As shown by the users' performance in our user study in Section \ref{subsection:results-user-study}, predicting the Most Replayed data is an arduous task. One plausible explanation is that the ground truth is noisy and lacks any clear patterns with respect to the input. Upon manual analysis of several videos, we found it hard to justify the location of certain peaks in the Most Replayed data.
Another possible explanation is that video-only input does not provide enough information to resolve the problem effectively. Some peaks may be caused by interesting information in the speech, while others could result from dramatic changes in loudness. We defer to further research to incorporate multimodal inputs, particularly from audio channels and text transcripts.

\section{Conclusion}
The Most Replayed data presents a new source of insight on users' interests in online video streaming.
In this work, we focus on predicting this data using Deep Learning and compare the results against human performance, measured through a user study. For our experiments, we use YTMR500, a novel dataset comprising 500 vlog videos and their corresponding Most Replayed data.
All the DL models evaluated on on YTMR500 perform significantly better than random, whereas the human participants are not able to accurately predict the Most Replayed data.
We believe future research can further enhance the results, for instance, by incorporating multimodal inputs. We encourage the community to deploy our dataset in follow-up work.



\null
\vskip .375in
\twocolumn[
\begin{center} 
  {\Large \bf Supplementary Material \par}
  \vspace*{24pt}
  {
  \large
  \lineskip .5em
  
  \par
  }
  \vskip .5em
  \vspace*{12pt}

\end{center}
]

\setcounter{equation}{0}
\setcounter{figure}{0}
\setcounter{table}{0}
\setcounter{section}{0}
\makeatletter
\renewcommand{\theequation}{S\arabic{equation}}
\renewcommand{\thefigure}{S\arabic{figure}}
\renewcommand{\bibnumfmt}[1]{[S#1]}
\renewcommand{\citenumfont}[1]{S#1}

\section{The YTMR500 dataset}
\label{section:supplementary_ytmr500}


Given a video, we investigate whether it is possible to predict which specific parts users will watch and replay. Since we are the first to tackle this problem, we cannot use pre-existing datasets. 
Therefore, we introduce a novel dataset of videos and annotations collected from YouTube, \emph{YTMR500} (``\textbf{Y}ou\textbf{T}ube \textbf{M}ost \textbf{R}eplayed \textbf{500}''). 
The dataset consists of 500 videos, in the form of with pre-computed spatio-temporal features, and the corresponding Most Replayed data. The videos have average duration of 11.9$\pm$4.1 minutes.
The dataset creation process can be outlined in three main steps: (1) Data collection, (2) Feature extraction and (3) Cleanup.

\vspace{-0.4cm}

\paragraph{Step 1a: Video data collection.}\label{para:ytmr50-step1a}
We retrieve videos from YouTube matching the following criteria:
\begin {enumerate*} [label=\itshape\roman*\upshape)]
\item under the Creative Commons license, \item duration from 3 to 20 minutes, and \item at least 30 thousand views. 
\end {enumerate*}
A high view count is necessary in order for YouTube to make the Most Replayed data publicly visible.
We use the search queries ``vlog'', ``trip'', ``travel'', ``visiting'' to obtain a list of videos, which are then selected to exclude those with static backgrounds or bad recording quality.
Some thumbnails of the videos in the dataset can be seen in Figure \ref{fig:thumbs}.

\vspace{-0.4cm}

\paragraph{Step 1b: Retrieval of the Most Replayed data.}
\label{para:ytmr50-step1b}
    For each of the videos in \hyperref[para:ytmr50-step1a]{Step 1a}, we download \footnote{https://github.com/Benjamin-Loison/YouTube-operational-API} the Most Replayed data, discarding those for which the data is missing.
    Concretely, the Most Replayed data consists of 100 scalars in the $[0,1]$ range, for each video. Hence, for videos of varying duration, each data point covers a different time duration.

\vspace{-0.4cm}

\paragraph{Step 2: Feature extraction}\label{para:ytmr50-step2}
    We use a pre-trained model to extract features from the videos. The model takes as input a segment of 32 frames, which corresponds to $\sim1.1$s of video, and outputs a feature vector of 1024 dimensions. 
    This helps reduce the complexity of the input to our trained model upfront and makes the architecture compatible with any resolution and video format. It also reduces the dataset size, with our settings, by a factor of 2.7 (from 7GB to 2.6GB).
    Multiple models were taken into consideration for this step, either image-based or video-based. 
    Image features, common in video summarization works \cite{PglSum, fajtl2019vasnet, Song2015TVSum}, cannot capture any temporal relationship between the frames of the segment.
    For this reason, we opt to use video features in our preprocessing.
    We choose \emph{I3D} as it is pre-trained on the extensive Kinetics-400~\cite{kay2017kinetics} and is a popular choice in the literature  \cite{doshi2022endtoend, girdhar_video_2019, Li2021BridgingTextAndVideo, mandal2019outofdistribution}. 
    Specifically, we obtain RGB features from the ``Mixed 5c'' layer of \emph{I3D}, \ie, the second-to-the-last layer, which are 1024-dimensional vectors \footnote{https://github.com/v-iashin/video\_features}.  
    Note that, for uniformity, all videos are first downsampled to 30 fps before feature extraction.

\vspace{-0.4cm}

\paragraph{Step 3: Cleanup}
    To reach the number of 500, a few excess videos are removed.
    The Most Replayed data annotations are transformed into a vector of size $100$, removing unnecessary metadata.
    The annotations are packaged together with the video features, having size $1024 \times T$, with $T$ proportional to video duration.
%

\section{Predicting the Most Replayed data in a video}
\label{section:supplementary_method}

\subsection{Deep Learning models} \label{section:model}

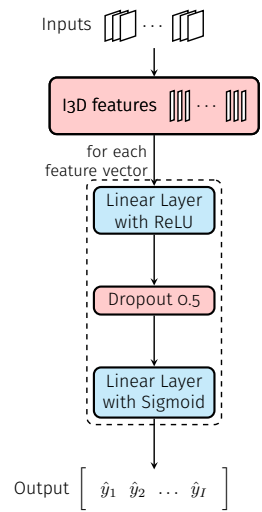
\begin{figure}[!ht]
    \centering
    \begin{adjustbox}{width=0.195\textwidth}
    \begin{tikzpicture}[font=\sffamily,module/.style={draw, very thick, rounded corners, minimum width=15ex},
smallmodule/.style={draw, very thick, rounded corners, minimum width=5ex},
embmodule/.style={module, fill=red!20},
segmodule/.style={smallmodule, fill=red!20},
gamodule/.style={module, fill=orange!20},
localmodule/.style={module, fill=orange!20},
mhamodule/.style={module, fill=orange!20},
lnmodule/.style={module, fill=yellow!20},
ffnmodule/.style={module, fill=cyan!20},
arrow/.style={-stealth, thick, rounded corners},
line/.style={thick, rounded corners},
]
\matrix [column sep=0cm, anchor=center] (feat) {
     \foreach \n [evaluate=\n as \x using (\n)*0.15] in {0,...,2} {
    \draw[fill=white!20,thick] (\x,-0.3) -- (\x+0.0,0.275) -- (\x+0.075,0.3) -- (\x+0.075,-0.275) -- cycle;
      }; &
    \node[align=center] {$\dots$}; &
    \foreach \n [evaluate=\n as \x using (\n)*0.15] in {0,...,2} {
        \draw[fill=white!20,thick] (\x,-0.3) -- (\x+0.0,0.275) -- (\x+0.075,0.3) -- (\x+0.075,-0.275) -- cycle;
    };\\
};
\node[left=0cm of feat] (i3dtext) {I3D features};
\begin{scope}[on background layer]
\node[fit={(feat)(i3dtext)}, embmodule, align=center] (i3d) {};
\end{scope}
\matrix [above=of i3d, column sep=0cm, anchor=center] (frames) {
 1 &
 \foreach \n [evaluate=\n as \x using (\n)*0.15] in {0,...,2} {
\draw[fill=white!20,thick] (\x,-0.3) -- (\x+0.0,0.2) -- (\x+0.3,0.3) -- (\x+0.3,-0.2) -- cycle;
  }; &
\node[align=center] {$\dots$}; &
\foreach \n [evaluate=\n as \x using (\n)*0.15] in {0,...,2} {
    \draw[fill=white!20,thick] (\x,-0.3) -- (\x+0.0,0.2) -- (\x+0.3,0.3) -- (\x+0.3,-0.2) -- cycle;
};\\
};

\node[left=0cm of feat] (i3dtext) {I3D features};
\begin{scope}[on background layer]
\node[fit={(feat)(i3dtext)}, embmodule, align=center] (i3d) {};
\end{scope}

\node[left=0cm of frames] (inputstext) {Inputs};


\begin{scope}[on background layer]
\node[fit={(feat)(i3dtext)}, segmodule, align=center] (i3d) {};
\end{scope}

\node[below=of i3d, ffnmodule, align=center] (fc1) {Linear Layer\\with ReLU};
\node[below=of fc1, embmodule, align=center] (do2) {Dropout 0.5};
\node[below=of do2, ffnmodule, align=center] (fc2) {Linear Layer\\with Sigmoid};

\matrix[below=of fc2] (array) [matrix of math nodes, left delimiter={[}, right delimiter={]}] {
    \hat{y}_{1} & \hat{y}_{2} & \dots & \hat{y}_{I} \\
};
\node[left=0.25cm of array] (outputstext) {Output};

\node[fit={(fc1)(do2)(fc2)}, draw, dashed, thick, rounded corners] (baseline) {};

\draw[arrow] (frames) -- (i3d);

\draw[arrow] (i3d) -- node[left, text width=2.3cm, align=right, font=\sffamily\small]{for each\\ feature vector} (fc1);

\draw[arrow] (fc1) -- (do2); 
\draw[arrow] (do2) -- (fc2);
\draw[arrow] (fc2) -- (array);





\end{tikzpicture}
    \end{adjustbox}
    \caption{Our fully-connected baseline. We use a $50\%$ Dropout to prevent the model from overfitting the training set.}
    \label{fig:model-linear}
\end{figure}

\begin{figure*}[!h] 
\centering
\includegraphics[width=1\textwidth]{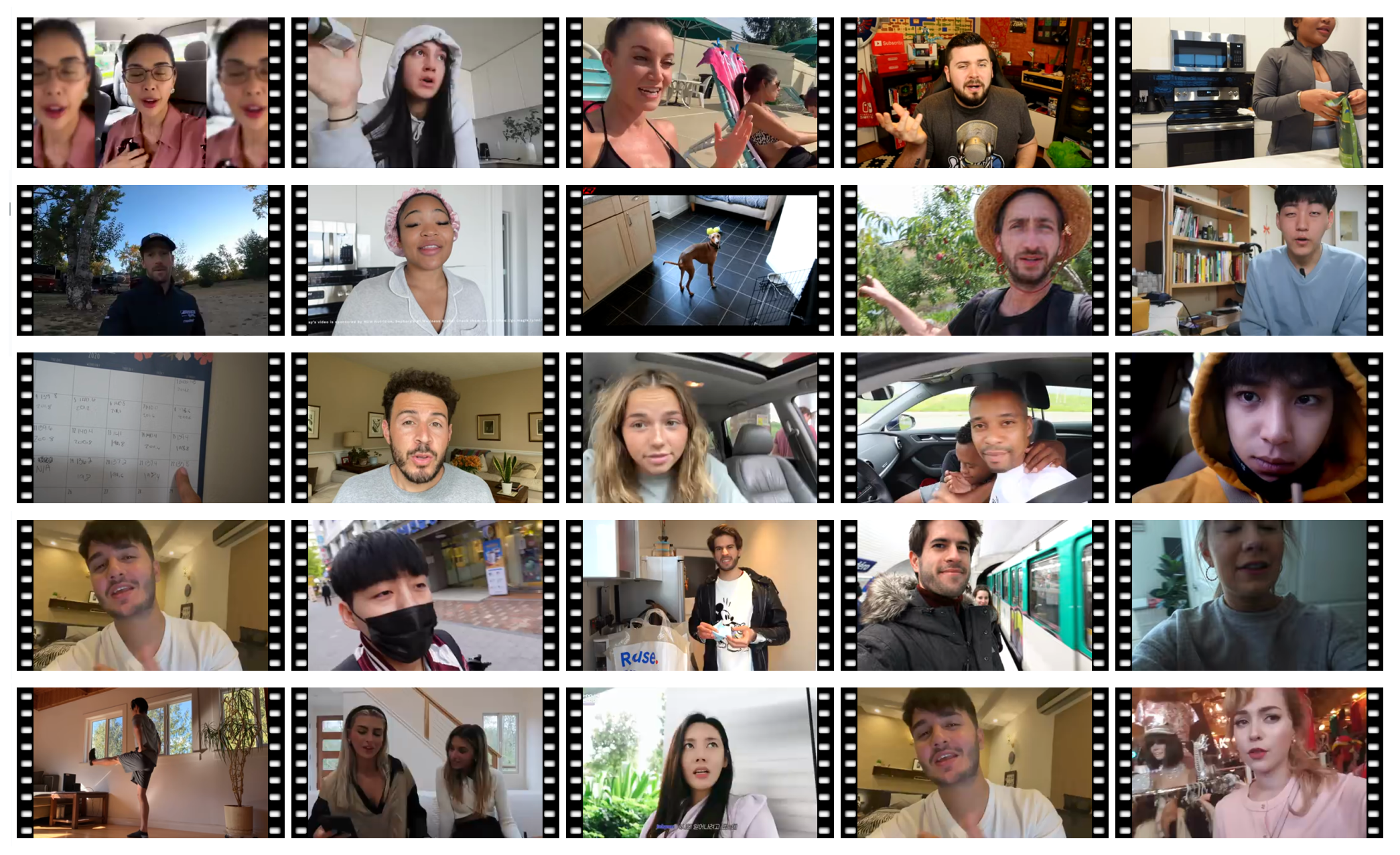}
\caption{
We introduce the \emph{YTMR500} dataset, which contains 500 vlog and travel videos and the corresponding Most Replayed data.}
\label{fig:thumbs}
\end{figure*}

\paragraph{Fully connected model}
We deploy as baseline a fully-connected model, shown in Figure \ref{fig:model-linear}. The model maps each input feature vector to a scalar in the range $[0,1]$.
This architecture does not capture any temporal relationships between different segments of the video.

\vspace{-0.4cm}

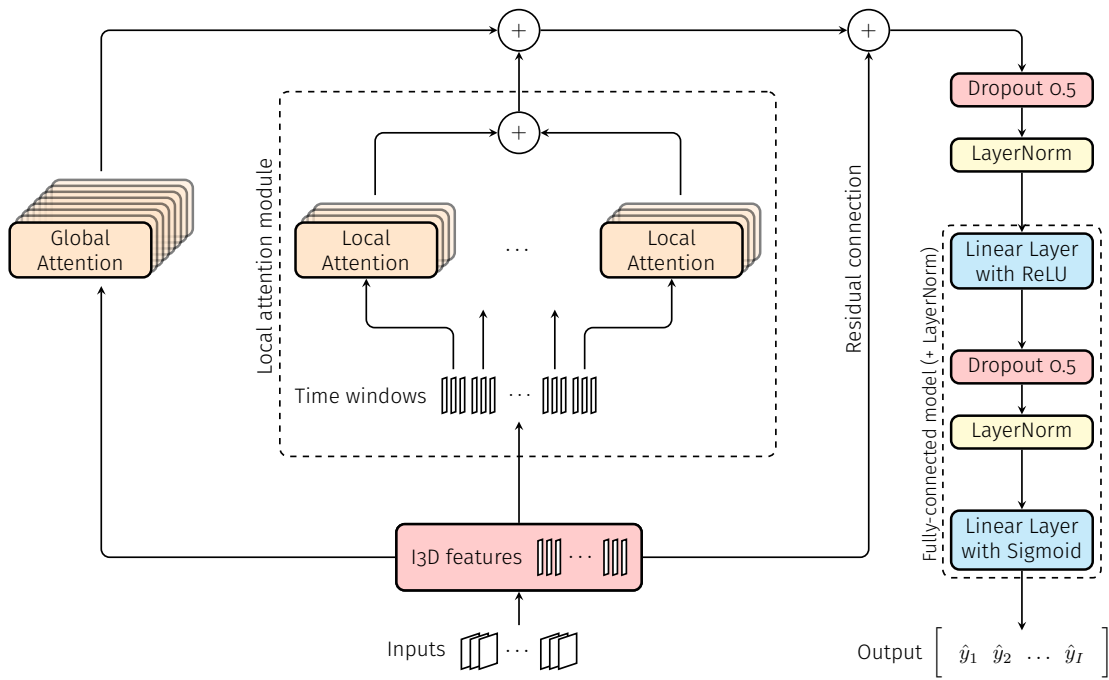
\begin{figure*}[!h]
    \centering
    \begin{adjustbox}{width=0.85\textwidth}
    \begin{tikzpicture}[font=\sffamily,module/.style={draw, very thick, rounded corners, minimum width=15ex},
smallmodule/.style={draw, very thick, rounded corners, minimum width=5ex},
embmodule/.style={module, fill=red!20},
segmodule/.style={smallmodule, fill=red!20},
gamodule/.style={module, fill=orange!20},
localmodule/.style={module, fill=orange!20},
mhamodule/.style={module, fill=orange!20},
lnmodule/.style={module, fill=yellow!20},
ffnmodule/.style={module, fill=cyan!20},
arrow/.style={-stealth, thick, rounded corners},
line/.style={thick, rounded corners},
]
\matrix [column sep=0cm, anchor=center] (feat) {
     \foreach \n [evaluate=\n as \x using (\n)*0.15] in {0,...,2} {
    \draw[fill=white!20,thick] (\x,-0.3) -- (\x+0.0,0.275) -- (\x+0.075,0.3) -- (\x+0.075,-0.275) -- cycle;
      }; &
    \node[align=center] {$\dots$}; &
    \foreach \n [evaluate=\n as \x using (\n)*0.15] in {0,...,2} {
        \draw[fill=white!20,thick] (\x,-0.3) -- (\x+0.0,0.275) -- (\x+0.075,0.3) -- (\x+0.075,-0.275) -- cycle;
    };\\
};
\node[left=0cm of feat] (i3dtext) {I3D features};
\begin{scope}[on background layer]
    \node[fit={(feat)(i3dtext)}, embmodule, align=center] (i3d) {};
\end{scope}
\matrix [below=of i3d, column sep=0cm, anchor=center] (frames) {
 1 &
 \foreach \n [evaluate=\n as \x using (\n)*0.15] in {0,...,2} {
\draw[fill=white!20,thick] (\x,-0.3) -- (\x+0.0,0.2) -- (\x+0.3,0.3) -- (\x+0.3,-0.2) -- cycle;
  }; &
\node[align=center] {$\dots$}; &
\foreach \n [evaluate=\n as \x using (\n)*0.15] in {0,...,2} {
    \draw[fill=white!20,thick] (\x,-0.3) -- (\x+0.0,0.2) -- (\x+0.3,0.3) -- (\x+0.3,-0.2) -- cycle;
};\\
};
\node[left=0cm of frames] (inputstext) {Inputs};

\matrix[above=1.7cm of i3d, column sep={0.05cm, 0cm, 0cm, 0.05cm}] (seg) {
\foreach \n [evaluate=\n as \x using (\n)*0.15] in {0,...,2} {
    \draw[fill=white!20,thick] (\x,-0.3) -- (\x+0.0,0.275) -- (\x+0.075,0.3) -- (\x+0.075,-0.275) -- cycle;
      }; &
      \foreach \n [evaluate=\n as \x using (\n)*0.15] in {0,...,2} {
    \draw[fill=white!20,thick] (\x,-0.3) -- (\x+0.0,0.275) -- (\x+0.075,0.3) -- (\x+0.075,-0.275) -- cycle;
      }; &
      \node[align=center, text width=10px, anchor=center] {\small$\dots$}; &
      \foreach \n [evaluate=\n as \x using (\n)*0.15] in {0,...,2} {    \draw[fill=white!20,thick] (\x,-0.3) -- (\x+0.0,0.275) -- (\x+0.075,0.3) -- (\x+0.075,-0.275) -- cycle;
      }; &
      \foreach \n [evaluate=\n as \x using (\n)*0.15] in {0,...,2} {
    \draw[fill=white!20,thick] (\x,-0.3) -- (\x+0.0,0.275) -- (\x+0.075,0.3) -- (\x+0.075,-0.275) -- cycle;
      }; \\
};

\begin{scope}[on background layer]
\node[fit={(feat)(i3dtext)}, segmodule, align=center] (i3d) {};
\end{scope}

\node[left=0cm of seg] (segtext) {Time windows};

\matrix [above=2cm of seg, column sep=1cm,anchor=center] (local) {
    \node[localmodule, text width=2cm, align=center] (local1) {Local Attention};  &
    \node[align=center] {$\dots$}; &
    \node[localmodule, text width=2cm, align=center] (localn) {Local Attention}; \\
  };

\node[above=of local, draw, thick, circle] (localplus) {$+$};
\node[above=of localplus, draw, thick, circle] (plus) {$+$};

\coordinate[right=5.5cm of plus] (residualhelper);
\coordinate (segfork) at ($(seg.north)!0.5!(local.south)$);
\coordinate (arrow1) at ($(i3d)!0.666!(seg)$); 
\coordinate (arrow2) at ($(localplus)!0.333!(plus)$);
\coordinate (arrowtip); 

\node[draw, thick, circle] (resplus) at (residualhelper) {$+$};
\node[left=2.25cm of local, gamodule, align=center] (ga) {Global\\Attention};

\begin{scope}[on background layer]
    \foreach \n [evaluate=\n as \x using (\n)*0.1] in {7,...,1} {
        \node[gamodule, opacity=0.5, minimum width=2.365cm, minimum height=0.95cm] (ga-fade\n) at ([xshift=\x cm, yshift=\x cm] ga.center) {};
    };

    \foreach \n [evaluate=\n as \x using (\n)*0.1] in {3,...,1} {
        \node[gamodule, opacity=0.5, minimum width=2.365cm, minimum height=0.95cm] (local1-fade\n) at ([xshift=\x cm, yshift=\x cm] local1.center) {};
    };
    \foreach \n [evaluate=\n as \x using (\n)*0.1] in {3,...,1} {
        \node[gamodule, opacity=0.5, minimum width=2.365cm, minimum height=0.95cm] (localn-fade\n) at ([xshift=\x cm, yshift=\x cm] localn.center) {};
    };


\end{scope}

\node[fit={(ga-fade1) (ga-fade2) (ga-fade3)(ga-fade4)(ga-fade5)(ga-fade6)(ga-fade7)(ga)}] (ga-full) {};
\node[fit={(local1-fade1) (local1-fade2) (local1-fade3)(local1)}] (local1-full) {};
\node[fit={(localn-fade1) (localn-fade2) (localn-fade3)(localn)}] (localn-full) {};

\node[fit={(local)(localn-full)(local1-full)(localplus)(seg)(segtext)(arrow1)(arrow2)}, draw, dashed, thick, rounded corners] (localnode) {};
\node[left=of localnode.north west, rotate=90, xshift=-1cm, yshift=-0.75cm] (textlocalmodule) {Local attention module};


\node[right=of resplus, yshift=-1cm, embmodule, align=center] (do1) {Dropout 0.5};
\node[below=0.5cm of do1, lnmodule, align=center] (ln1) {LayerNorm};
\node[below=of ln1, ffnmodule, align=center] (fc1) {Linear Layer\\with ReLU};
\node[below=of fc1, embmodule, align=center] (do2) {Dropout 0.5};
\node[below=0.5cm of do2, lnmodule, align=center] (ln2) {LayerNorm};
\node[below=of ln2, ffnmodule, align=center] (fc2) {Linear Layer\\with Sigmoid};

\matrix[below=of fc2] (array) [matrix of math nodes, left delimiter={[}, right delimiter={]}] {
    \hat{y}_{1} & \hat{y}_{2} & \dots & \hat{y}_{I} \\
};

\node[left=0.25cm of array] (outputstext) {Output};

\node[fit={(fc1)(do2)(ln2)(fc2)}, draw, dashed, thick, rounded corners] (baseline) {};
\node[left=of baseline.north west, rotate=90, yshift=-0.8cm, xshift=-0.25cm, font=\sffamily\small] (fctext) {Fully-connected model (+ LayerNorm)};

\draw[arrow] (frames) -- (i3d);
\draw[arrow] ([xshift=-1.1cm]seg.north) |- ([xshift=-1.5cm]segfork) -|  (local1);
\draw[arrow] ([xshift=-0.6cm]seg.north) -- ([xshift=-0.6cm, yshift=1cm] seg.north);
\draw[arrow] ([xshift=0.6cm] seg.north) -- ([xshift=0.6cm, yshift=1cm] seg.north); 
\draw[arrow] ([xshift=1.1cm]seg.north) |- ([xshift=1.5cm]segfork)   -| (localn);
\draw[arrow] (plus) -- (resplus);
\draw[arrow] (i3d) -- (seg);
\draw[arrow,] (i3d) -| (ga-full);
\draw[arrow] (localplus) -- (plus);
\draw[arrow] (local1-full) |- (localplus);
\draw[arrow] (localn-full) |- (localplus);
\draw[arrow] (ga-full.north) |- (plus);
\draw[arrow] (i3d) -| node[above,sloped,xshift=5cm]{Residual connection} (resplus);

\draw[arrow] (resplus) -| (do1);
\draw[arrow] (do1) -- (ln1);
\draw[arrow] (ln1) -- (fc1);
\draw[arrow] (fc1) -- (do2); 
\draw[arrow] (do2) -- (ln2);
\draw[arrow] (ln2) -- (fc2);
\draw[arrow] (fc2) -- (array);





\end{tikzpicture}
    \end{adjustbox}
    \caption{Our attention-based model architecture is inspired by PGL-SUM of Apostolidis \etal~\cite{PglSum}. The Global Attention module can capture relationships between each of the segments of the video, while the Local Attention modules focus on relationships inside a section. Ultimately, two linear layers map the vector output of the attention modules into a scalar, for each segment.}
    \label{fig:model-attention}
\end{figure*}

\paragraph{Attention-based model} \label{para:model-attention}
We investigate whether considering the temporal relationships across segments improves the MR prediction accuracy beyond our fully-connected baseline. 
For this, we use an attention-based model inspired by the architecture of PGL-SUM (Apostolidis \etal~\cite{PglSum}).
The model is shown in Figure \ref{fig:model-attention}.

\clearpage

\section*{Acknowledgements}
\noindent This work is part of the research program Efficient Deep Learning (EDL), which is (partly) financed by the Dutch Research Council (NWO).

{\small
\bibliographystyle{ieee_fullname}
\bibliography{egbib,egbib_youtube}
}

\end{document}